\documentclass[final,3p,twocolumn,sort&compress]{elsarticle}

\usepackage{lineno,hyperref}
	\hypersetup{colorlinks=true,pdfborder=0 0 0}
\usepackage{subcaption}
\usepackage{amsmath}
\usepackage{amssymb}
\usepackage{gensymb}
\usepackage{cleveref}
\usepackage{siunitx}
\usepackage[dvipsnames]{xcolor}
\usepackage{xspace}
\modulolinenumbers[5]
\usepackage{graphicx}
\usepackage{booktabs}

\usepackage{fancyhdr}

\usepackage{etoolbox}
\makeatletter
\patchcmd{\ps@pprintTitle}
  {Preprint submitted}
  {Postprint of article submitted}
  {}{}
\makeatother

\journal{Acta Materialia}

\begin{document}

\begin{frontmatter}

\title{Machine learning of microstructure--property relationships in materials leveraging microstructure representation from foundational vision transformers}


\address[az-am]{Graduate Interdisciplinary Program in Applied Mathematics, University of Arizona, Tucson, AZ 85721, USA}
\address[az-mse]{Department of Materials Science and Engineering, University of Arizona, Tucson, AZ 85721, USA}

\author[az-am]{Sheila E. Whitman}
\author[az-am,az-mse]{Marat I. Latypov\corref{cor1}}
\cortext[cor1]{corresponding author}
\ead{latmarat@arizona.edu}

\begin{abstract}
Machine learning of microstructure--property relationships from data is an emerging approach in computational materials science. Most existing machine learning efforts focus on the development of task-specific models for each microstructure--property relationship. We propose utilizing pre-trained foundational vision transformers for the extraction of task-agnostic microstructure features and subsequent light-weight machine learning of a microstructure-dependent property. We demonstrate our approach with pre-trained state-of-the-art vision transformers (CLIP, DINOv2, SAM) in two case studies on machine-learning: (i) elastic modulus of two-phase microstructures based on simulations data; and (ii) Vicker's hardness of Ni-base and Co-base superalloys based on experimental data published in literature. Our results show the potential of foundational vision transformers for robust microstructure representation and efficient machine learning of microstructure--property relationships without the need for expensive task-specific training or fine-tuning of bespoke deep learning models. \\

\noindent Note: \textcolor{red}{this is an author-generated postprint} of the article by Whitman \& Latypov \href{https://doi.org/10.1016/j.actamat.2025.121217}{published} in {\it{Acta Mater.}}. \\ DOI: 10.1016/j.actamat.2025.121217

\end{abstract}


\end{frontmatter}

\pagestyle{fancy}
\fancyhf{}
\fancyhead[LO]{Postprint of \href{https://doi.org/10.1016/j.actamat.2025.121217}{Whitman \& Latypov, Acta Mater (2025) 121217}}

\section{Introduction}
\label{sec:intro}

Structural alloys represent an important class of materials needed across all critical industries (energy, defense, transportation, infrastructure). Design of structural alloys relies on quantitative understanding of microstructure--property relationships. Computer models capable of capturing these relationships can significantly accelerate materials design endeavors. Machine learning is rapidly emerging as a powerful computational tool with models successfully trained on experiments \cite{khatavkar2020accelerated,mahdavi2023reduced}, physics-based simulations \cite{latypov2019materials,ibragimova2022convolutional,hu2024anisognn,sim2025fip}, or their combinations \cite{pagan2022graph}.

Enabling machine learning of microstructure--property relationships in structural materials relies on quantitative description of the microstructure. Robust description of microstructure is a non-trivial task because of the rich diversity of microstructures observable at different length scales and a variety of their aspects (spatial, geometric, statistical) relevant for properties \cite{kalidindi2015hierarchical}. One strategy is to use geometric descriptors of microstructures (e.g., phase volume fraction, grain size) that are intuitive and familiar from traditional models (e.g., Voigt/Reuss bounds, Hall--Petch relation \cite{hall1954variation,petch1956xvi}). Another strategy is to describe microstructures with distribution functions: $n$-point correlations \cite{jiao2007modeling,Adams1998, mahdavi2023reduced}, lineal path functions \cite{lu1992lineal}, or chord length distributions \cite{torquato1993chord,latypov2018application,whitman2024sr}. This strategy was shown successful for modeling a range of properties based on data from both  experiments (e.g., \cite{khatavkar2020accelerated,mahdavi2023reduced}) and simulations (e.g, \cite{latypov2019materials,latypov2017data}). A limitation of machine learning with traditional microstructure descriptors is the need to select the most appropriate set of descriptors or distribution functions for each individual property-specific model \cite{xu2015machine}. Besides geometric and statistical microstructure descriptions inspired by micromechanics theories, purely data-driven approaches (e.g., CNNs) have also been explored. CNNs for modeling microstructure--property relationships are typically designed and trained from scratch for each specific property of interest \cite{cecen2018material,yang2018deep,ibragimova2022convolutional,pokharel2021physics}. However, training task-specific CNNs and designing their architectures for a variety of microstructure--property relationships is data-intensive, time-consuming, and computationally expensive. 

While most machine learning studies on structural materials focus on task-specific models, research on language modeling and computer vision has undergone a paradigm shift towards task-agnostic foundational models \cite{he2020momentum}. Foundational models learn representations of high-dimensional data (texts, images) that are advantageously universal for a spectrum of downstream tasks. Modeling with universal features can yield even better results than task-specific neural networks \cite{brown2020language}. This progress has been possible with the advent of the transformer architecture \cite{vaswani2017attention} and strategies for unsupervised learning from large unlabeled datasets \cite{caron2018deep,he2022masked,chen2020simple}. SAM, CLIP, and DINOv2 are examples of recently developed foundational models in the field of computer vision. All of these models produce rich feature representations of images with a semantic meaning but differ in their unique specialty and pre-training strategy. CLIP focuses on learning multi-modal representations of images and the corresponding captions by maximizing their cosine similarity \cite{radford2021learning}. SAM allows promptable segmentation through a training process involving both manual and automated mask annotation \cite{kirillov2023segment}, and DINOv2 utilizes discriminative self-supervised learning between image-level and patch-level features to create task-agnostic representations of images \cite{oquab2023dinov2}. Given the success of these models on unseen computer vision tasks, materials research could benefit from the adoption and development of foundational models that facilitate learning relationships without task-specific reinvention of architectures, expensive training, or fine-tuning. 

In this study, we demonstrate and evaluate multiple pre-trained vision transformers (ViTs) as microstructure feature extractors for machine learning of microstructure--property relationships. We hypothesize that the general-purpose visual features that pre-trained ViTs extract from images can serve as robust microstructure representation for modeling properties without training or fine-tuning the ViTs to any materials data. Using features obtained with the ViTs, we train simple regression-type models that predict engineering properties from the microstructure. In this paper, we first describe our approach in detail (\Cref{sec:methods}) and then evaluate its application in two case studies (\Cref{sec:results}): elastic stiffness of synthetic two-phase microstructures learned from simulation data (\Cref{sec:sim}) and microhardness of Ni-base and Co-base superalloys learned from experimental data (\Cref{sec:exp}). We additionally present the incorporation of compositional data as additional features besides microstructure in representation of the superalloys in \Cref{sec:chem}.

\section{ViT approach to modeling microstructure--property relationships}
\label{sec:methods}

Our proposed approach (illustrated in \Cref{fig:exp,fig:approach}) utilizes microstructure features from images obtained with pre-trained ViTs for material property prediction. This ViT-based approach involves the following steps: 
\begin{enumerate}
    \item collect training data: microstructure images and their corresponding properties of interest;
    \item obtain image-level features with a pre-trained ViT by a ``forward pass'' of each microstructure image through the transformer;
    \item aggregate features from multiple images if multiple images are available for the same microstructure;
    \item reduce the dimensionality of high-dimensional feature vectors;
    \item train a lightweight regression-type machine learning model that captures the relationship between microstructure features and property. 
\end{enumerate}

\begin{figure}[!h]
  \centering
   \includegraphics[width=0.45\textwidth]{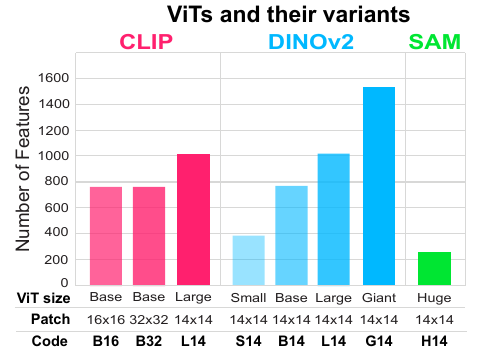}
   \caption{ViTs and their variants tested in this study with the number of features that each ViT generates for a single image. Throughout this paper, we refer to different variants by a letter and an number, with letter designating the ViT size and number specifying the patch size (see "Code"). For example, ``CLIP B16'' designates the base size of CLIP with a patch size of $16\times16$ pixels.}
   \label{fig:models}
\end{figure}

In this work, we test and critically evaluate three state-of-the-art ViTs and their variants (\Cref{fig:models}): three CLIP variants (base and large with different patch sizes) \cite{radford2021learning}, four DINOv2 variants (small, base, large, and giant with the same patch size) \cite{oquab2023dinov2} and one SAM variant (huge) \cite{kirillov2023segment}. ViTs process images in patches --- an elementary unit of the image similar to tokens in language processing \cite{vaswani2017attention, dosovitskiy2020image}. The patch size depends on the ViT and its variant: $14\times14$ pixels for SAM and DINOv2; $14\times14$, $16\times16$, or $32\times32$ pixels for different CLIP variants. Depending on the ViT and the raw microstructure data, step \#2 may require pre-processing of the images to make them compatible with the size and format expected by each ViT. Specifically, all ViTs expect RGB images; in addition, SAM and CLIP models require specific image sizes ($224\times224$ and $1024\times1024$, respectively), while  DINOv2 only requires the width and height of the input images to be multiples of the patch size. Therefore, pre-processing would typically involve conversion to the RGB format, resizing, and/or cropping (see pre-processing applied to specific microstructure data in \Cref{sec:sim,sec:exp}). 

For appropriately formatted images, a forward pass through a pre-trained ViT with fixed weights produces the desired microstructure features. CLIP and DINOv2 output a multidimensional image-level token called ``classification'' (or [CLS]) token \cite{radford2021learning,oquab2023dinov2}. Since the [CLS] token summarizes the visual information about the entire image, we directly adopt it as a feature vector representing the microstructure for machine-learning microstructure--property relationships. The output of SAM models includes only patch-level tokens and does not contain the image-level [CLS] token. Therefore, representing microstructures with SAM models requires an additional step of aggregating the patch-level features. The dimensionality of patch- and image-level tokens depends on the architecture of the ViTs and is specifically dictated by the size of the final hidden layer used for image encoding. Since the ViTs and their variants used in this study have different architectures (including differing hidden layers), they produce microstructure features of varying "lengths", as shown in \Cref{fig:models}. 

If available, multiple images (e.g., orthogonal or oblique 2D sections) from the same microstructure may be individually passed through a ViT followed by aggregation of their features into a single microstructure feature vector. In this work, we explore concatenation and element-wise mean pooling of vectors as two aggregation methods. Depending on the ViT and its size, the feature vectors may be large and contain more than \SI{1000}{} elements (\Cref{fig:models}). To focus on the most salient features, enable efficient machine learning, and avoid overfitting, we reduce the dimensionality of the extracted microstructure descriptors as part of the overall approach. Different techniques (e.g., UMAP \cite{mcinnes2018umap}, or t-SNE \cite{van2008visualizing}) can be used; here we adopt principal component analysis (PCA)
given its successful use with high-dimensional statistical descriptions \cite{latypov2017data,latypov2019materials,niezgoda2013novel}. Following PCA, we train simple machine learning models (linear, polynomial, support vector machines) using regression to obtain a quantitative relationship between a property of interest and the reduced-order representation of the microstructure. 

\section{Results}
\label{sec:results}

Here, we present the results of using the proposed ViT framework for learning and predicting the microstructure dependence of elastic stiffness of two-phase materials and Vicker's hardness (HV) from experimental data on Ni-base and Co-base superalloys. For both case studies, we compare simple regression-type models trained on microstructure features (i) obtained with ViTs (as proposed in \Cref{sec:methods}); (ii) obtained with a domain-specific CNN \cite{stuckner2022microstructure}; and (iii) represented by two-point correlations \cite{latypov2017data,hu2022learning}. 

\subsection{Case study 1: Young’s modulus of two-phase material (simulations)}
\label{sec:sim}
 
Our first case study focuses on machine-learning Young's modulus of 3D two-phase microstructures. To this end, we leverage a published dataset of \SI{5900}{} two-phase 3D microstructures and their corresponding overall modulus values obtained with finite element simulations \cite{cecen2018material}. The microstructures represented  by binary voxel data consist of a stiff phase and a compliant phase with a stiffness ratio of 50 -- a relatively high property contrast, which is generally challenging for traditional models \cite{latypov2017data,yang2018deep}.

\begin{figure*}[ht]
  \centering
   \includegraphics[width=\textwidth]{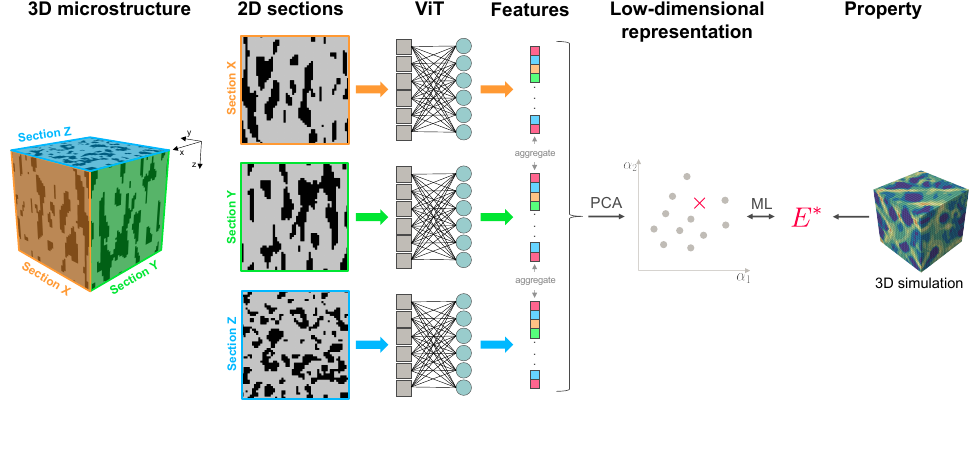}
   \caption{Machine learning of effective Young's modulus ($E^\ast$) of two-phase materials using ViT-based microstructure description and aggregation of features from multiple 2D sections of 3D microstructures pursued in Case Study 1. The procedure is illustrated for one sample (out of \SI{5900}{}); the simulation data is from Cecen et al.\ \cite{cecen2018material}.}
   \label{fig:approach}
\end{figure*}

In this case study, we aimed to predict the overall Young's modulus from three orthogonal 2D sections of the microstructure (\Cref{fig:approach}). First, 2D microstructure images are much more widely accessible than 3D data given the high cost and need in highly specialized and expensive equipment for 3D characterization \cite{echlin2012new,pokharel2015situ,uchic2007three}. Second, 2D microstructure images are readily compatible with pre-trained ViTs, which typically work with 2D images or photographs in the general, non-materials domain of computer vision. Finally, property prediction based on three orthogonal 2D sections of microstructure was recently shown feasible with non-ViT microstructure descriptions \cite{hu2022learning}.

Having three orthogonal sections for each microstructure, we first obtained features for each individual section (\Cref{fig:approach}). To make the sections compatible with input to the ViTs, the binary images were resized and then converted to RGB. Resizing depended on the ViT as SAM and CLIP expect specific image sizes, while DINOv2 is more flexible and only requires the image width and height to be multiples of the patch size. Therefore, for DINOv2 with a patch size of $14\times14$, each $51\times51$ section (binary image of 0 and 1 pixels) was cropped to size $42\times42$ as 42 is a multiple of 14 closest to 51. For SAM and CLIP requiring $224\times224$ and $1024\times1024$ images as input, we split each pixel in the microstructure into a small patch ($5\times5$ for SAM and $21\times21$ for CLIP) and assigned the phase label of the original pixel to all of the new pixels occupying the same location. We then cropped the resulting upsampled ($255\times255$ and $1071\times1071$) images by selecting the top-left region to match the input sizes expected by the two ViTs. This resizing strategy, suitable for binary images, avoids artifacts and only results in a minor loss of pixels at the bottom-right edges of the microstructure. 

\begin{figure*}[ht]
  \centering
   \includegraphics[width=\linewidth]{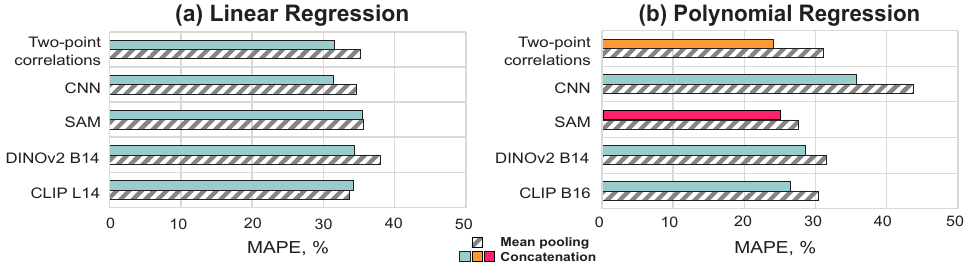}
   \caption{Accuracy of Young's modulus predictions for the test set shown in terms of MAPE for (a) linear models and (b) second-order polynomial models obtained by regression using ViT features, two-point correlations, and domain-specific CNN features.}
   \label{fig:dig_mape}
\end{figure*}

In addition to ViT features, we calculated spatial correlations and obtained features with a CNN fine-tuned to microstructure data. We calculated two-point autocorrelations for the stiff phase using the Spatial Correlation Toolbox implemented in MATLAB \cite{Cecen2015}. Two-point autocorrelations calculated for $51\times51$ microstructure images resulted in $51\times51$ probability maps, subsequently reshaped into \SI{2601}{}-element feature vectors. As a domain-specific CNN, i.e., a CNN ``familiar'' with microstructures of materials, we adopted a CNN developed for classification of micrographs of multiphase alloys trained on \SI{110861}{} microstructure images ("MicroNet" dataset) \cite{stuckner2022microstructure}. To obtain microstructure representation with this CNN, we passed the images through the network with fixed weights trained on the MicroNet dataset except the final classification layer. The output of the network with the ResNet50 architecture without the classification layer resulted in \SI{2048}{} values that we used as microstructure features for machine learning. 

For all three descriptions (ViT features, two-point correlations, and domain-specific CNN features) we aggregated the three feature vectors from the three sections of each microstructure using either concatenation or mean pooling (see \Cref{fig:approach}). Following aggregation, we carried out PCA for dimensionality reduction of the aggregated features. We treated the number of principal components for machine learning as a tunable hyperparameter. 

For training, hyperparameter tuning, and testing, we held out \SI{10}{\percent} of the dataset as a test set and split the remaining \SI{5310}{}  samples into training and validation subsets with an $80:20$ ratio. We trained and compared linear regression (LR) and second-order polynomial regression models (PR) on the training set and used the mean absolute percentage error (MAPE) for the validation set as the error metric to minimize when searching for the optimal number of principal components. Since the number of principal components was our only hyperparameter, we used grid search as the hyperparameter tuning strategy. 

\begin{figure*}[ht]
  \centering
   \includegraphics[width=0.81\linewidth]{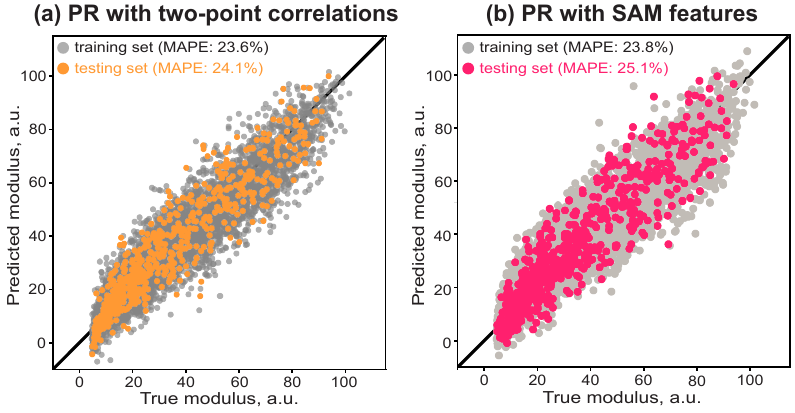}
   \caption{Prediction of Young's modulus shown as parity plots for a set of 590 unseen microstructures by second-order polynomial regression trained on (a) 24 principal components of concatenated two-point correlations and the (b) 39 principal components of concatenated SAM features}. 
   \label{fig:dig_par}
\end{figure*}

\Cref{fig:dig_mape} presents the results of the LR and PR models predicting Young's modulus for the test set unseen during training. The results are shown for the cases of concatenation and mean pooling of features obtained with the ViTs, domain-specific CNN, and two-point correlation calculations. Here we visualize only the best performing variants of CLIP and DINOv2, however, all ViT variants shown in \Cref{fig:models} were tested. In almost all cases, concatenation of features for the three 2D sections leads to more accurate regression models compared to aggregation by mean pooling. With the  exception of the domain-specific CNN features, the PR models (MAPE below \SI{30}{\percent} for most cases using concatenation) outperformed all corresponding LR models (MAPE above \SI{30}{\percent}) indicating a nonlinear relationship between the overall Young's modulus and the microstructure. Among the studied cases, the lowest MAPE of \SI{24.1}{\percent} is obtained with a PR model that uses 24 principal components of two-point correlations. PR models trained on the concatenated features obtained with SAM achieve a slightly higher MAPE value of \SI{25.1}{\percent}. \Cref{fig:dig_par} shows parity plots comparing ground truth values of the Young's modulus with those from the best PR models based on two-point correlations and SAM features for the training and testing sets.

\begin{figure*}[ht]
  \centering
   \includegraphics[width=\linewidth]{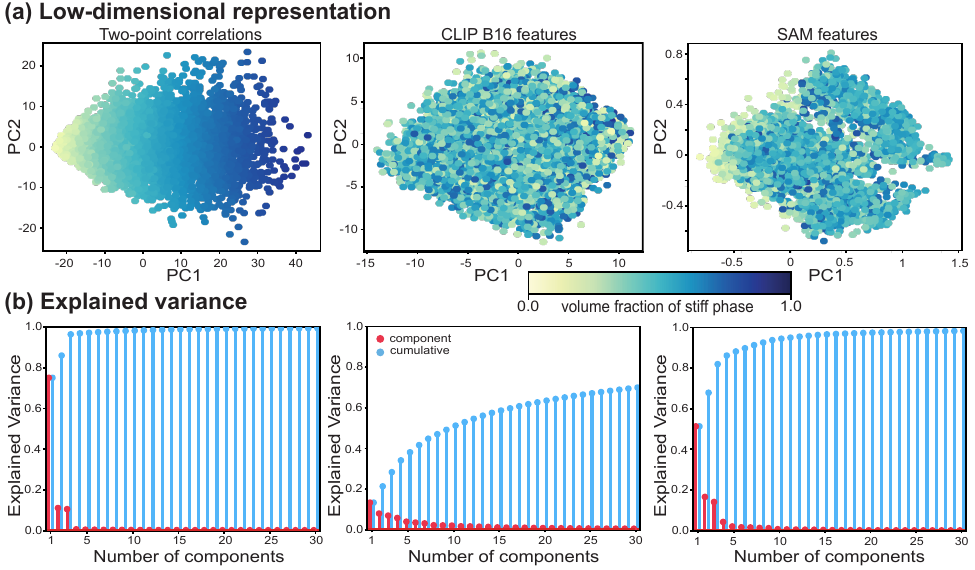}
   \caption{PCA of microstructure features representing 
   \SI{5900}{} 2D sections of two-phase microstructures: (a) low-dimensional representation in terms of the first two principal components 
   and (b) explained variance by the first 30 principal components (cumulative and per component).}
   \label{fig:dig_pca}
\end{figure*}

To better understand the microstructure features obtained with the ViTs in this case study, we visualize their low-dimensional representation in terms of the first two principal components from PCA (\Cref{fig:dig_pca}). We focus on the PCA of the two best performing sets of features -- obtained with CLIP and SAM -- and compare it with PCA of two-point correlations previously discussed in literature \cite{niezgoda2013novel,latypov2017data}. PCA leads to dense low-dimensional representations without pronounced clusters in all cases. The most striking difference between the two representations is that the first principal component of the two-point correlations is highly correlated with the volume fraction of the stiff phase, which is not the case for the principal components of the ViT features. Indeed, the volume fraction steadily increases along the horizontal axis from zero to one (represented by color in \Cref{fig:dig_pca}a). The first principal component of the two-point correlation function (highly correlated with the volume fraction of the stiff phase) is also a significantly dominant one, capturing \SI{75}{\percent} variance in the dataset as seen in the scree plot (\Cref{fig:dig_pca}b). At the same time, the first principal component of SAM features explains about \SI{50}{\percent} and there is even a smaller gap in the variance explained by the first few principal components in the case of the CLIP features (\Cref{fig:dig_pca}b). 

\subsection{Case study 2: Vicker's hardness of superalloys (experiments)}
\label{sec:exp}

\begin{figure*}[ht]
  \centering
   \includegraphics[width=\textwidth]{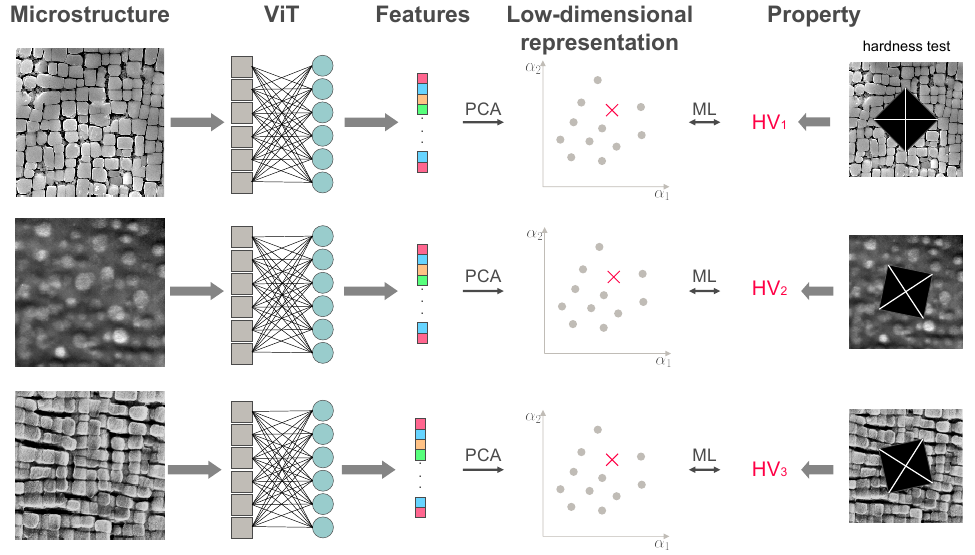}
   \caption{Machine learning of Vicker's hardness (HV) of superalloys using ViT-based microstructure description of grayscale micrographs pursued in Case Study 2. The procedure is illustrated for three representative samples (out of 149) collected from literature. The hardness test is depicted schematically only. The three images are from \cite{peng2018microstructure, picasso2009nucleation, zhang2018effects}.} 
   \label{fig:exp}
\end{figure*}

In the second case study, we utilized the ViT framework to predict Vicker's hardness (HV) of Ni-base and Co-base superalloys from their microstructures based on experimental measurements (\Cref{fig:exp}). To this end, we extracted 149 scanning electron microscopy images (SEM) and their corresponding hardness from 19 papers, similar to a recent study using two-point correlations as the microstructure description \cite{khatavkar2020accelerated}. Most hardness values in the 19 papers are reported in \SI{}{\kilo\gram f \per\milli\meter\squared} units. We converted the remaining data (28 values) in GPa, to the consistent units using the relationship \SI{1}{\kilo\gram f \per\milli\meter\squared} $=10^3/g$ \SI{}{\giga\pascal} with $g$ denoting the standard gravity \cite{astm2017standard}. 

\begin{figure*}[ht]
  \centering
   \includegraphics[width=\textwidth]{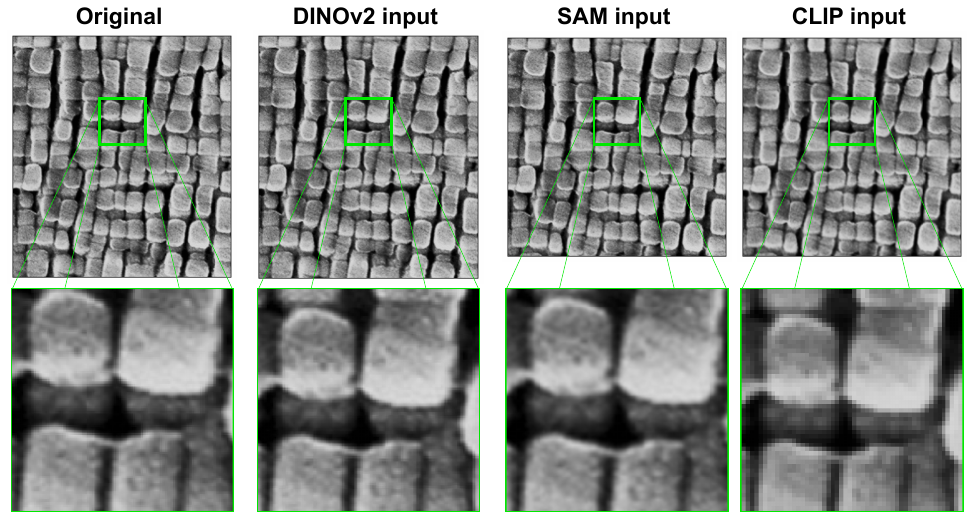}
   \caption{Pre-processing of experimental microstructure images for input to the three ViTs exemplified by one of the microstructures (raw image after \cite{zhang2018effects}). }
   \label{fig:interp}
\end{figure*}

As in the first case study, the experimental images (now grayscale unlike binary in \Cref{sec:sim})  were pre-processed for the ViTs. Pre-processing an experimental image of an arbitrary size included cropping, conversion to RGB, and resizing (for CLIP and SAM only). For DINOv2, which does not require a specific input size, the images were cropped such that both the width and height were the largest multiples of the patch size ($14\times14$): for example, a raw image of $662\times731$ ("original" in \Cref{fig:interp}) was cropped to $658\times728$ ("DinoV2 input" in \Cref{fig:interp}). For CLIP and SAM, which require specific input sizes, images were cropped to the largest square whose side length is a multiple of the corresponding ViT's patch size. The cropped images were then converted to RGB and either upscaled or downscaled to $224\times224$ for CLIP, and upscaled to  $1024\times1024$ for SAM. Bilinear interpolation was used for both downscaling and upscaling, implemented in PyTorch as the \texttt{resize} function \cite{paszke2017automatic}. As a specific example of resizing, a raw $662\times731$ image (shown in \Cref{fig:interp}) was cropped to $656\times656$ and downscaled to $224\times224$ for CLIP. For SAM, the same raw image was cropped to $658\times658$ and then upscaled to $1024\times1024$. The post-processing results obtained for each ViT, exemplified by a representative image, show that the microstructure was largely preserved during the process without significant artifacts (\Cref{fig:interp}). Following pre-processing, the images were passed through the ViTs to obtain the microstructure feature vectors.

As benchmark representations, we once again calculated the two-point correlations and obtained features from the domain-specific CNN (\cite{stuckner2022microstructure}) for comparison with ViT features. Since the experimental images are grayscale (unlike binary in the first case study), calculation of the two-point correlations of the phases required an additional step of image segmentation. Segmentation is necessary to clearly identify regions occupied by each phase. Only in segmented images can two-point correlations be clearly defined and calculated as probabilities of pairs of points in a phase, or a combination of phases in the case of cross-correlations. To segment the images, we utilized the following workflow (adapted from Ref.~\cite{khatavkar2020accelerated}) with the aid of the OpenCV package in Python \cite{opencv_library}: (i) convert images from RGB to BGR (expected input to OpenCV functions), (ii) denoise images with a non-local denoising method, and (iii) segment with adaptive thresholding. Following segmentation, we calculated two-point cross-correlation functions for the resulting binary images using the Spatial Correlation Toolbox \cite{Cecen2015}. We focused on cross-correlations in this case study (unlike autocorrelations in \Cref{sec:sim}) because these functions describe probabilities of finding a matrix and a precipitate at any pair of pixels in the microstructure (within the cut-off radius \cite{cecen2016versatile}) {\it{independent}} of whether the precipitates or the matrix appears as the dark/light phase in any given image of the diverse dataset. For further consistency in cross-correlation maps of differing size obtained for microstructure images of various sizes, we center-cropped the correlation maps to a size $159\times159$ corresponding to the smallest microstructure map in our dataset. Finally, we reshaped each $159\times159$ probability map to a \SI{25281}{}-element feature vector.

With all three types of features, we used PCA for dimensionality reduction and trained three classes of models using LR, PR, and support vector regression (SVR) to obtain the microstructure dependence of microhardness. We additionally tested the SVR model in this case study due to its suitability for machine learning based on small datasets \cite{smola2004tutorial,mesut2023role}. For the same reasons of limited data (149 samples), we used cross validation to perform hyperparameter tuning and evaluate the performance of the machine learning models \cite{james2023introduction}. We utilized nested 10-fold cross-validation with grid search to first select the optimal number of principal components followed by re-training and evaluation of the LR, PR, and SVR models. 

\begin{figure*}[ht]
  \centering
   \includegraphics[width=\linewidth]{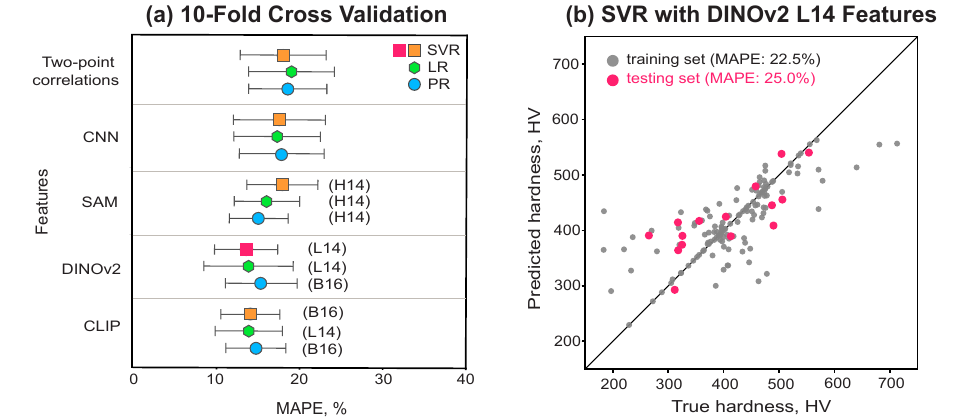}
   \caption{Machine learning of Vicker's hardness based on microstructure features shown in a (a) mean and standard deviation error plot with a (b) parity plot of results obtained with the best performing model --- an SVR trained on 34 principal components of microstructure features from  DINOv2 L14.}
   \label{fig:exp_mape}
\end{figure*}

\Cref{fig:exp_mape} visualizes the 10-fold cross validation results for the LR, PR, and SVR models using ViT features, domain-specific CNN features, and two-point correlations. The results are shown in terms of the mean and standard deviation of MAPE values obtained across different folds. As in the first case study, we visualize the results only for a single, most accurate variant of both CLIP and DINOv2. The SVR model using 34 principal components of the microstructure feature vector obtained with DINOv2 L14 leads to the lowest mean MAPE; the parity plot for this model is shown in \Cref{fig:exp_mape}b. Converse to the results in the first case study, the models based on two-point correlations had the highest mean MAPE in all three regression cases.  

\subsection{Complementing microstructure description with composition information}
\label{sec:chem}

Building on the work of Khatavkar et al.\ \cite{khatavkar2020accelerated}, we explored improving property predictions by introducing alloy compositions into the model input in addition to the microstructure representations. For our dataset of Ni- and Co-base superalloys, concentrations of 22 elements constitute the compositions. For each set of the microstructure feature vectors (ViT, domain-specific CNN, two-point correlations), we appended the corresponding 22-element composition vectors to form an enhanced alloy representation as input for machine learning models. Following concatenation of the microstructure features and the elemental compositions, we standardized all the combined vectors to have a zero mean and a unit standard deviation. We then carried out PCA and trained the three regression models (LR, PR, and SVR) to capture the dependence of microhardness on both microstructure and composition of the superalloys. 

\begin{figure*}[ht]
  \centering
   \includegraphics[width=\linewidth]{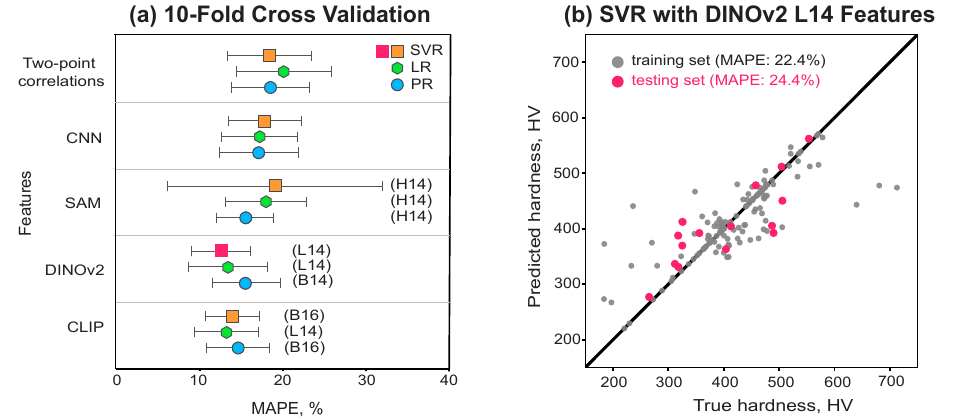}
   \caption{Machine learning of Vicker's hardness based on microstructure and composition shown in a (a) mean and standard deviation error plot with a (b) parity plot of results obtained with the best performing model --- an SVR trained on 49 principal components of the DINOv2 L14 features concatenated with elemental composition vectors.}
   \label{fig:exp_mape_comp}
\end{figure*}

\Cref{fig:exp_mape_comp} shows the 10-fold cross validation results for the regression models using the new concatenated feature vectors that include both compositions and microstructure features (ViT, domain-specific CNN, and two-point correlations). Overall, the results with the addition of the compositions are similar to those obtained with the microstructure description as the only input (\Cref{fig:exp_mape}). However, some models show slight improvements: e.g., SVR based on DINOv2 L14 features improves mean MAPE by \SI{.9}{\percent}. This SVR model, which uses 49 principal components of DINOv2 L14 features, shows the best overall accuracy in our cross validation (\Cref{fig:exp_mape_comp}). Interestingly, the addition of the compositional information to SAM features results in a significantly large standard deviation of MAPE for an SVR model compared to all other cases (\Cref{fig:exp_mape_comp}a). The discrepancy in this particular case is caused by an outlier MAPE of \SI{54.69}{\percent} obtained for one of the folds. This fold includes an alloy (from \cite{chen2024microstructure}) with a distinct composition having significantly higher concentrations of boron and silicon compared to all other alloys in the dataset: \SI{2.8}{\percent} vs.\ less than \SI{0.1}{\percent} for boron and \SI{3.5}{\percent} vs.\ below \SI{0.1}{\percent} for silicon. Without this outlier, the standard deviation for the same machine learning model drops from \SI{12.93}{\percent} to \SI{5.42}{\percent}, comparable to all other regression results. 

\section{Discussion}
\label{sec:disco}

The two case studies presented above tested our hypothesis that foundational ViTs trained on very large datasets of general (non-materials) images can serve as microstructure feature extractors for machine-learning microstructure--property relationships in alloys. Polynomial models of Young's modulus in two-phase alloys trained on simulations data had comparable accuracy (\SI{1}{\percent} difference in MAPE)  when based on best-performing ViT features and two-point correlations as the microstructure representation. 

At the same time, ViT features served as a better microstructure description for machine-learning microhardness as a function of microstructure from experimental data (\Cref{fig:exp_mape}). We attribute this distinct outcome 
to the difference in the raw microstructure images in the two datasets: the simulation dataset contained binary images, whereas the experimental dataset consisted of grayscale images, which require segmentation as an additional step for machine learning based on two-point correlations. Segmentation is required to clearly distinguish the constituent phases for defining and computing physically meaningful two-point correlation functions. Indeed, it is the two-point correlations for phases as discrete microstructure states that serve as statistical description of their spatial configuration and thus fundamentally determine properties of multiphase materials \cite{torquato2002random,gupta2015structure,latypov2019materials}. Yet, segmentation of real-world experimental images can be non-trivial, dependent on imaging conditions, and prone to errors \cite{bales2017segmentation}. Segmentation errors negatively impact the calculation accuracy of two-point statistics (as other geometric descriptors \cite{whitman2024automated}) and the corresponding machine learning models. In contrast, ViTs can provide features for non-discrete images without the need for segmentation. The second case study showed the advantage of avoiding the segmentation step and the associated errors: all machine learning models using ViT features outperformed the same models based on two-point correlations (\Cref{fig:exp_mape}). Interestingly, complementing ViT features with compositional information only marginally improved the machine learning models of the microhardness for superalloys (\Cref{fig:exp_mape_comp} vs.\ \Cref{fig:exp_mape}). One interpretation of this result is that the microstructure implicitly ``encodes'' compositional effects and that the microstructure alone might be sufficiently predictive of such properties as Vicker's hardness without explicit account for the composition. However, whether this finding is specific to the dataset and its limitations (size, diversity) or universal for a wide range of materials and properties needs further investigation. 

In addition to better accuracy for real-world images and simpler workflows without segmentation, machine learning based on ViT features offer additional benefits of (i) modest requirements to the size of training datasets, and (ii) computational efficiency, when compared to training or even fine-tuning task-specific deep learning models. The pre-trained ViTs considered in this work provide microstructure features ``out of the box'': that is, without training or fine-tuning to any materials-specific data. Trained on very large datasets of natural images, the ViTs learned universal features, providing the benefits of a transformer model without the need for large domain-specific training datasets. This is especially advantageous for materials science applications with scarcely available training data. Without the need to train a task-specific CNN or fine-tune a ViT, the approach studied here is computationally efficient. For the larger dataset of \SI{5900}{} microstructures, extracting features from the three 2D cross-sections using DINOv2 or CLIP takes between \SI{7}{\minute} (smaller models) and \SI{7}{\hour} (larger models) on a consumer-grade laptop (MacBook Air M1 with \SI{16}{\giga\byte} RAM). Training a task-specific CNN model on the same dataset from scratch was reported to take \SI{48}{\hour} on a K80 GPU \cite{cecen2018material}. Only SAM ViT feature extraction requires a comparable \SI{47}{\hour} (although without a GPU) due to the large input image size of $1024\times1024$. 

We note that the microstructure representation and property inference could be further improved with fine-tuning ViTs to microstructure data. Generally, fine-tuning a base deep learning model to a specific downstream task is a computationally efficient strategy with modest training data requirements compared to training from scratch. This strategy has been effective with CNNs for addressing such materials problems as microstructure segmentation \cite{stuckner2022microstructure} or learning microstructure--property relationships \cite{xu2021method}. With ViTs, however, even fine-tuning comes at a much higher computational cost because of the much larger number of parameters (than CNNs) in state-of-the-art ViTs and quadratic complexity of self-attention \cite{devoto2024adaptive}. Whether improvements in accuracy from fine-tuning ViTs to a specific microstructure ensemble justify the requirements in terms of the computational resources needs further investigation.

While microstructure features obtained with ViTs or their reduced order representation (from PCA) do not lend themselves a trivial interpretation (as opposed to, e.g., spatial correlations), we gained insight by comparing principal components of ViT features with those of two-point correlations (\Cref{fig:dig_pca}). PCA of two-point correlations for microstructures with a wide range of phase volume fractions often leads to a largely dominant first principal component that highly correlates with the volume fraction of the phase for which the two-point autocorrelation is calculated (see \Cref{fig:dig_pca}a and \cite{latypov2017data}). A principal component that captures a large extent of the data variance while mostly representing a phase volume fraction may overlook more subtle details of the microstructure such as phase morphology or its spatial configuration. Capturing these details is essential for microstructure-sensitive property models.  We found that the first principal component of ViT features was decoupled from the phase volume fraction and the first principal component captured less variance in the ensemble of \SI{5900}{} two-phase microstructures (\Cref{fig:dig_pca}b). These characteristics of reduced-order representation of microstructures using principal components of ViT features can serve as a basis of property models with high sensitivity to fine microstructure details. 

These results and findings in this study show the potential of machine learning approaches based on robust representations of microstructures independent of the specific material class or specific target properties. The development of materials-focused, yet foundational, ViTs (or other deep learning architectures) could prove even more powerful for universal microstructure description.

\section{Conclusions}

In summary, we demonstrated the potential of foundational ViTs for feature extraction from microstructure images for supervised learning of microstructure--property relationships. The key idea of this approach is to use pre-trained ViTs to obtain robust microstructure descriptions without training or fine-tuning these ViTs (or any other bespoke deep learning models) for each microstructure dataset or property of interest. Our first case study of ViT features for machine-learning Young's modulus from simulation data led to the following conclusions: 

\begin{enumerate}
    \item The overall Young's modulus of two-phase materials can be predicted from microstructure features obtained and aggregated from three orthogonal 2D sections with about \SI{25}{\percent} error on average for microstructures unseen during training.
    \item Concatenation of feature vectors from three orthogonal sections consistently gives better accuracy than aggregation by calculating the element-wise mean of the feature vectors.
    \item Among features obtained with three pre-trained ViTs, 
    SAM features result in the lowest error on a test set (\SI{25.1}{\percent} MAPE), while two-point correlations as a microstructure description leads to a polynomial model with the best accuracy (\SI{24.1}{\percent} MAPE).
    \item The principal components of ViT features are more balanced in terms of explained variance than the principal components of two-point correlations, where the first component captures \SI{75}{\percent} of the variance in the dataset; the first principal component of ViT features is not as correlated with phase volume fractions as in the case of two-point correlations. 
\end{enumerate}

We further draw the following conclusions from the second case study on machine learning of Vicker's hardness of superalloys: 

\begin{enumerate}

\item Machine learning with ViT features leads to better accuracy than comparable models using two-point correlations in all considered scenarios. 
\item Unlike the calculation of two-point correlations for phases, microstructure description using ViT features eliminates the need for phase segmentation in experimental images avoiding negative impacts of segmentation errors on the accuracy of property models. 
\item An SVR model with DINOv2 features achieves the lowest 10-fold cross validation mean MAPE of \SI{13.5}{\percent} (vs.\ \SI{17.9}{\percent} MAPE obtained with two-point correlations and \SI{17.2}{\percent} MAPE with domain-specific CNN features).
\item Complementing microstructure descriptions with compositional information leads to overall similar results as machine learning with microstructure only; an SVR model with DINOv2 features appended with alloy compositions reaches a \SI{0.9}{\percent} improvement in terms of mean MAPE over the best result without compositional information.  

\end{enumerate}

\section*{Acknowledgments}

SEW acknowledges the support by the National Science Foundation Graduate Research Fellowship Program under Grant No.\ DGE-2137419. MIL acknowledges the support by the National Science Foundation under Award No.\ 2441813. The views and conclusions contained herein are those of the authors and should not be interpreted as necessarily representing the official policies or endorsements, either expressed or implied, of the National Science Foundation.

\section*{Code availability}
The code for this paper is made available on GitHub \url{https://github.com/materials-informatics-az/MicroPropViT}.

\section*{Conflict of interest}

The authors declare that there is no conflict of interest.

\bibliography{refs.bib}

\end{document}